%% file: main.tex
\documentclass{article} 
\usepackage{ijcai16,times}
\usepackage[english]{babel}
\usepackage{hyperref}
\usepackage{url}
\usepackage{alltt}
\usepackage{epsfig,endnotes}
\usepackage{url}
\usepackage{appendix}
\usepackage{subfigure}
\usepackage{xspace}
\usepackage{listings}
\usepackage[binary-units=true]{siunitx}
\usepackage{color,colortbl}
\definecolor{Gray}{gray}{0.925}
\definecolor{DarkGray}{gray}{0.8}
\usepackage{amssymb,amsmath}
\usepackage{amsthm}
\usepackage{textcomp}
\usepackage{float}
\newcommand{\pullWeights}{{\tt{pullWeights}}\xspace}
\newcommand{\pushGradient}{{\tt{pushGradient}}\xspace}
\newcommand{\getMinibatch}{{\tt{getMinibatch}}\xspace}
\newcommand{\calcGradient}{{\tt{calcGradient}}\xspace}
\newcommand{\applyUpdate}{{\tt{applyUpdate}}\xspace}
\newcommand{\sumGradients}{{\tt{sumGradients}}\xspace}
\newcommand{\ts}{\textsuperscript}
\newcommand{\CIFAR}{{\tt{CIFAR10}}\xspace}
\newcommand{\imagenet}{{\tt{ImageNet}}\xspace}
\newcommand{\minibatch}{{mini-batch}\xspace}

\newcommand{\assign}{:=}

\newcommand{\tmop}[1]{\ensuremath{\operatorname{#1}}}

\newtheorem{remark}{Remark}
\newtheorem{theorem}{Theorem}

\title{Staleness-aware Async-SGD for Distributed Deep Learning}

\author{Wei Zhang, Suyog Gupta \\
IBM T. J. Watson Research Center\\
Yorktown Heights, NY 10598, USA \\
\texttt{\{weiz,suyog\}@us.ibm.com} \\
\And
Xiangru Lian \& Ji Liu \\
Department of Computer Science \\
University of Rochester, NY 14627, USA \\
\texttt{\{lianxiangru,ji.liu.uwisc\}@gmail.com} \\
}

\newcommand{\wztext}[1]{{  #1 }}

\begin{document}

\maketitle

\begin{abstract}

Deep neural networks have been shown to achieve state-of-the-art performance in several machine learning tasks. Stochastic Gradient Descent (SGD) is the preferred optimization algorithm for training these networks and asynchronous SGD (ASGD) has been widely adopted for accelerating the training of large-scale deep networks in a distributed computing environment. However, in practice it is quite challenging to tune the training hyperparameters (such as learning rate) when using ASGD so as achieve convergence and linear speedup, since the stability of the optimization algorithm is strongly influenced by the asynchronous nature of parameter updates. In this paper, we propose a variant of the ASGD algorithm in which the learning rate is modulated according to the gradient staleness and provide theoretical guarantees for convergence of this algorithm. Experimental verification is performed on commonly-used image classification benchmarks: CIFAR10 and Imagenet to demonstrate the superior effectiveness of the proposed approach, compared to SSGD (Synchronous SGD) and the conventional ASGD algorithm. 
\end{abstract}

\input{intro.tex}

\input{sysarch.tex}

\input{theory_hook.tex}
\input{result.tex}

\input{conclusion.tex}

\bibliographystyle{named}
\bibliography{iclr2016_ijcai2016}
\newpage
\appendix

\end{document}

%% file: intro.tex
\section{Introduction}

Large-scale deep neural networks training is often constrained by
the available computational resources, motivating the development of
computing infrastructure designed specifically for accelerating this
workload. This includes distributing the training across several
commodity CPUs (\cite{distbelief},\cite{adam}),or using heterogeneous
computing platforms containing multiple GPUs per computing node
(\cite{seide20141},\cite{minwa},\cite{amazon}), or using a CPU-based HPC
cluster (\cite{2015Gupta}).

Synchronous SGD (SSGD) is the most straightforward distributed implementation of SGD in which the master simply splits the workload amongst the workers at every iteration. Through the use of barrier synchronization, the master ensures that the workers perform gradient computation using the identical set of model parameters. The workers are forced to wait for the slowest one at the end of every iteration.   
This synchronization cost deteriorates the scalability and runtime performance of the SSGD algorithm. 
Asynchronous SGD (ASGD) overcomes this drawback by removing any explicit synchronization amongst the workers. 
However, permitting this asynchronous behavior inevitably adds ``staleness" to the system wherein some of the workers compute gradients using model parameters that may be several gradient steps behind the most updated set of model parameters. 
Thus when fixing the number of iterations, ASGD-trained model tends to be much worse than SSGD-trained model. 
Further, there is no known principled approach for tuning learning rate in ASGD to effectively counter the effect of stale gradient updates.

Prior
theoretical work by \cite{tsitsiklis1986distributed} and
\cite{agarwal2011distributed} and recent work by \cite{liu-asgd-nips-2015} provide
theoretical guarantees for convergence of stochastic optimization
algorithms in the presence of stale gradient updates for \emph{convex} optimization and \emph{nonconvex} optimization, respectively. We find that adopting the
approach of scale-out deep learning using ASGD gives rise to
complex interdependencies between the training algorithm's
hyperparameters (such as learning rate, \minibatch size) and the
distributed implementation's design choices (such as synchronization
protocol, number of learners), ultimately impacting the neural
network's accuracy and the overall system's runtime performance. In
practice, achieving good model accuracy through distributed training
requires a careful selection of the training hyperparameters and
much of the prior work cited above lacks enough useful insight to help
guide this selection process.

The work presented in this paper intends to fill this void by
undertaking a study of the interplay between the different design
parameters encountered during distributed training of deep neural
networks. In particular, we focus our attention on understanding the
effect of stale gradient updates during distributed training and
developing principled approaches for mitigating these effects. To this
end, we introduce a variant of the ASGD algorithm in which we
keep track of the staleness associated with each gradient computation
and adjust the learning rate on a per-gradient basis by simply dividing
the learning rate by the staleness value. The implementation of this 
algorithm on a CPU-based HPC cluster with fast interconnect is shown to 
achieve a tight bound on the gradient staleness. We experimentally 
demonstrate the effectiveness of the proposed staleness-dependent learning 
rate scheme using commonly-used image classification benchmarks: CIFAR10 and Imagenet and show that this simple, yet effective technique is necessary for
achieving good model accuracy during distributed training. Further, we
build on the theoretical framework of \cite{liu-asgd-nips-2015} and
prove that the convergence rate of the staleness-aware ASGD
algorithm is consistent with SGD:
$\mathcal{O}\left(1/\sqrt{T}\right)$ where $T$ is the number of
gradient update steps.

Previously, \cite{ho2013more} presented a parameter server based
distributed learning system where the staleness in parameter updates
is bounded by forcing faster workers to wait for their slower
counterparts. Perhaps the most closely related prior work is that of
\cite{chan2014distributed} which presented a multi-GPU system for
distributed training of speech CNNs and acknowledge the need to
modulate the learning rate in the presence of stale gradients. The
authors proposed an exponential penalty for stale gradients and show
results for up to 5 learners, without providing any theoretical guarantee of the convergence rate. However, in larger-scale distributed
systems, the gradient staleness can assume values up to a few hundreds
(\cite{distbelief}) and the exponential penalty may reduce the
learning rate to an arbitrarily small value, potentially slowing down the
convergence. In contrast, in this paper, we formally prove our proposed ASGD algorithm to converge as fast as SSGD. Further, our implementation achieves near-linear speedup while maintaining the optimal model accuracy. We demonstrate this on widely used image classification benchmarks.

%% file: sysarch.tex

\section{System architecture}
\label{sec:sysarch}
In this section we present an overview of our distributed deep learning system and describe the synchronization protocol design. In particular, we introduce the $n$-softsync protocol which enables a fine-grained control over the upper bound on the gradient staleness in the system. For a complete comparison, we also implemented the \textit{Hardsync protocol} (aka SSGD) for model accuracy baseline since it generates the most accurate model (when fixing the number of training epochs), albeit at the cost of poor runtime performance.


\subsection{Architecture Overview}
\label{sec:arch_overview}
We implement a parameter server based distributed learning system, which is a superset of Downpour SGD in \cite{distbelief}, to evaluate the effectiveness of our proposed staleness-dependent learning rate modulation technique. Throughout the paper, we use the following definitions:
\begin{itemize}
\item $\lambda$: number of learners (workers).
\item $\mu$: mini-batch size used by each learner to produce stochastic gradients.
\item $\alpha$: learning rate.
\item Epoch: a pass through the entire training dataset.
\item Timestamp: we use a scalar clock to represent weights timestamp $i$, starting from $i=0$. Each weight update increments the timestamp by 1. The timestamp of a gradient is the same as the timestamp of the weight used to compute the gradient.
\item $\tau_{i,l}$: staleness of the gradient from learner $l$.  A learner $l$ pushes gradient with timestamp $j$ to the parameter server of timestamp $i$, where $i \geq j$.We calculate the staleness $\tau_{i,l}$ of this gradient as $i-j$. $\tau_{i,l}\geq 0$ for any $i$ and $l$.
\end{itemize}
Each learner performs the following sequence of steps.
\getMinibatch : Randomly select a \minibatch of examples from the training data;   \pullWeights : A learner pulls the current set of weights from the parameter server; \calcGradient : Compute stochastic gradients for the current \minibatch. We divide the gradients by the \minibatch size; \pushGradient : Send the computed gradients to the parameter server;

The parameter server group maintains a global view of the neural network weights and performs the following functions.
\sumGradients : Receive and accumulate the gradients from the learners; \applyUpdate : Multiply the average of accumulated gradient by the learning rate (step length) and update the weights.

\subsection{Synchronization protocols}
\label{sec:sync_protocol}
\wztext{We implemented two synchronization protocols: hardsync protocol (aka, SSGD) and $n$-softsync protocol (aka, ASGD). 
Although running at a very slow speed, hardsync protocol provides the best model accuracy baseline number, when fixing the number of training epochs. $n$-softsync protocol is our proposed ASGD algorithm that automatically tunes learning rate based on gradient staleness and achieves model accuracy comparable with SSGD while providing a near-linear speedup in runtime.}

\textit{Hardsync protocol}: To advance the weights' timestamp $\theta$  from $i$ to $i+1$, each learner $l$ compute a gradient $\Delta\theta_{l}$ using a \minibatch size of $\mu$ and sends it to the parameter server. 
The parameter server averages the gradients over $\lambda$ learners and updates the weights according to equation~\ref{eqn:update_hard}, then broadcasts the new weights to all learners. The learners are forced to wait for the updated weights until the parameter server has received the gradient contribution from \emph{all} the learners and finished updating the weights. This protocol guarantees that each learner computes gradients on the exactly the same set of weights and ensures that the gradient staleness is 0. The hardsync protocol serves as the baseline, since from the perspective of SGD optimization it is equivalent to SGD using batch size $\mu\lambda$.
\begin{equation}
\begin{aligned}
	g_{i} &= \frac{1}{\lambda}\sum_{l=1}^{\lambda}{\Delta\theta_{l}} \\
	\theta_{i+1} &= \theta_{i} - \alpha g_{i}.
\end{aligned}
 \label{eqn:update_hard}
\end{equation}

\textit{$n$-softsync protocol}: Each learner $l$ pulls the weights from the parameter server, calculates the gradients and pushes the gradients to the parameter server. 
The parameter server updates the weights after collecting at least $c = \lfloor(\lambda / n) \rfloor$ gradients from \emph{any} of the $\lambda$ learners. Unlike hardsync, there are no explicit synchronization barriers imposed by the parameter server and the learners work asynchronously and independently.
The splitting parameter $n$ can vary from 1 to $\lambda$.
The $n$-softsync weight update rule is given by:
\begin{equation}
\begin{aligned}
	c &= \lfloor(\lambda / n) \rfloor \\
	g_{i} &= \frac{1}{c}{\sum_{l=1}^c \alpha({\tau_{i,l}}) \Delta\theta_{l}},~l \in \{1, 2,\hdots, \lambda\} \\
	\theta_{i+1} &= \theta_{i} - g_{i},
 \label{eqn:update_soft}
 \end{aligned}
\end{equation}
where $\alpha\left({\tau_{i,l}}\right)$ is the gradient staleness-dependent learning rate. \textit{Note that our proposed $n$-softsync protocol is a superset of Downpour-SGD of \cite{distbelief}(a commonly-used ASGD implementation), in that when $n$ is set to be $\lambda$, our implementation is equivalent to Downpour-SGD. By setting different $n$, ASGD can have different degrees of staleness, as demonstrated in Section~\ref{sec:staleness}.}

\subsection{Implementation Details}
\label{sec:impl_details}
We use MPI as the communication mechanism between learners and parameter servers. Parameter servers are sharded. Each learner and parameter server are 4-way threaded. During the training process, a learner pulls weights from the parameter server, starts training when the weights arrive, and then calculates gradients. Finally it pushes the gradients back to the parameter server before it can pull the weights again.  We do not ``accrue'' gradients at the learner so that each gradient pushed to the parameter server is always calculated out of  one \minibatch size as accruing gradients generally lead to a worse model. In addition, the parameter server communicates with learners via MPI blocking-send calls (i.e., \pullWeights and \pushGradient), that is the computation on the learner is stalled until the corresponding blocking send call is finished. The design choice is due to the fact that it is difficult to guarantee making progress for MPI non-blocking calls and multi-thread level support to MPI communication is known not to scale \cite{mpi-spec}. Further, by using MPI blocking calls, the gradients' staleness can be effectively bounded, as we demonstrate in Section~\ref{sec:staleness}. Note that the computation in parameter servers and learners are however concurrent (except for the learner that is communicating with the server, if any). No synchronization is required between learners and no synchronization is required between parameter server shards. 

Since memory is abundant on each computing node, our implementation does not split the neural network model across multiple nodes (model parallelism). Rather, depending on the problem size, we pack either 4 or 6 learners on each computing node. Learners operate on homogeneous processors and run at similar speed. In addition, fast interconnect expedites pushing gradients and pulling weights. Both of these hardware aspects help bound gradients' staleness.

\input{staleness}

%% file: staleness.tex
\subsection{Staleness analysis}
\label{sec:staleness}
\begin{figure}[t]
\centering
	\includegraphics[width=\columnwidth]		  
        {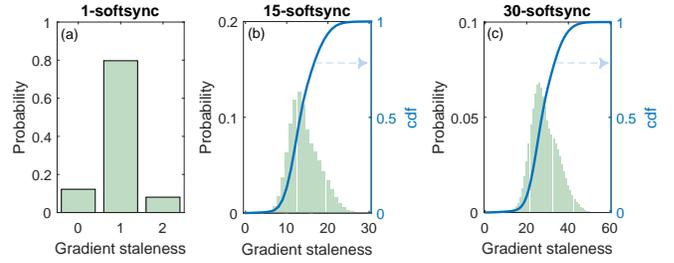}
\vskip -0.1in
	\caption{Distribution of gradient staleness for $1$, $15$, and $30$-softsync protocols. $\lambda = 30$}
	\label{fig:staleness}
\end{figure}
In the hardsync protocol, the update of weights from $\theta_i$ to $\theta_{i+1}$ is computed by aggregating the gradients calculated using weights $\theta_i$. 
As a result, each of the gradients $g_i$ in the $i\ts{th}$ step carries with it a staleness $\tau_{i,l}$ equal to 0. 

Figure~\ref{fig:staleness} shows the measured distribution of gradient staleness for different $n$-softsync (ASGD) protocols when using $\lambda=30$ learners.
For the $1$-softsync, the parameter server updates the current set of weights when it has received a total of 30 gradients from (any of) the learners. 
In this case, the staleness $\tau_{i,l}$ for the gradients computed by the learner $l$ takes values 0, 1, or 2. 
Similarly, the $15$-softsync protocol forces the parameter server to accumulate $\lambda/15=2$ gradient contributions from the learners before updating the weights. On the other hand, the parameter server updates the weights after receiving a gradient from any of the learners when the $30$-softsync protocol is enforced. The average staleness $\langle\tau_{i}\rangle$ for the $15$-softsync and $30$-softsync protocols remains close to 15 and 30, respectively. Empirically, we have found that a large fraction of the gradients have staleness close to $n$, and only with a very low probability ($<0.0001$) does $\tau$ exceed $2n$. These measurements show that, in general,  $\tau_{i,l} \in \{0,1,\hdots,2n\}$ and $\langle\tau_{i}\rangle \approx n$ for the $n$-softsync protocol \footnote{We have found this empirical observation to hold true regardless of the \minibatch size per learner and the size of the model. The plots in Figure~\ref{fig:staleness} were generated using the \CIFAR dataset/model (see section \ref{sec:results}) and \minibatch size per learner $\mu=4$.}. Clearly, the $n$-softsync protocol provides an effective mechanism for controlling the staleness of the gradients in the system.

In our implementation, the parameter server uses the staleness information to modulate the learning rate on a per-gradient basis. For an incoming gradient with staleness 
$\tau_{i,l}$, the learning rate is set as:
\begin{equation}
\label{eq:LR}
\alpha_{i,l} = \frac{\alpha_0}{\tau_{i,l}} \quad \text{if}~ \tau_{i,l} > 0
\end{equation}
where $\alpha_0$ is typically set as the `best-known' learning rate when using SSGD. In section \ref{sec:results} we show experimental results comparing this staleness-dependent learning rate scheme with the case where the learning rate is kept constant at $\alpha_0$.

%% file: theory_hook.tex
\section{Theoretical Analysis}
This section provides theoretical analysis of the ASGD
algorithm proposed in section~\ref{sec:sysarch}. More specifically, we will show the convergence rate and how the gradient staleness affects the convergence. In essence, we are solving the following generic optimization problem:
\[ \min_\theta \quad F (\theta) \assign \frac{1}{N} \sum_{i = 1}^N f_i
  (\theta), \] where $\theta$ is the parameter vector we are pursuing,
$N$ is the number of samples, and $f_i(\theta)$ is the loss function
for the $i^{\text{th}}$ sample. If every learner computes $\mu$
gradients at once and the parameter server updates the parameter when
it receives $c$ mini-batches from learners, from the perspective of
the parameter server, the update procedure of parameter $\theta$
mentioned can be written as
\begin{eqnarray}
  g_i & = & \underbrace{\frac{1}{c}  \sum_{l = 1}^c \frac{\alpha_0}{\tau_{i,
           l}}  \underbrace{\left( \frac{1}{{\mu}}  \sum_{s = 1}^{{\mu}} \nabla
           f_{\xi_{i, s, l}} (\theta_{i - \tau_{i, l}}) \right)}_{\text{calculated in
           a learner}}}_{\text{aggregated in the parameter server}},  \label{dl-eq:1}\\
  \theta_{i + 1} & = & \underbrace{\theta_i - g_i}_{\text{calculated in the
                      parameter server}} , \nonumber
\end{eqnarray}
where $\tau_{i, l}$ is the staleness of the parameter used to calculate
the gradients in the $l^{\text{th}}$ learner in step $i$ and
$\alpha_0$ is a constant. $\xi_{i, s, l}$ denotes the subscript of $f$
used to calculate the $s^{\tmop{th}}$ gradient in the $l^{\text{th}}$
learner in the $i^{\tmop{th}}$ step.

To simplify the analysis, we decompose every single step in
\eqref{dl-eq:1} to $c$ steps by updating only one batch of gradients
in each step. Then the sequence $\{\theta_i\}_{i\in\mathbb{N}}$
becomes $\{\tilde{\theta}_t\}_{t\in\mathbb{N}}$, and
$\tilde{\theta}_{ci}=\theta_i$. Formally, it will be
\begin{eqnarray}
  \tilde{g}_t & = & \frac{1}{c{\mu}}  \frac{\alpha_0}{\tilde{\tau}_t - r_t} 
  \underbrace{ \left(\sum_{s = 1}^{{\mu}} \nabla f_{\tilde{\xi}_{t, s}} (\tilde{\theta}_{t -
           \tau_t})\right)}_{\text{calculated in a learner}} , 
  \label{dl-eq:2}\\
  \tilde{\theta}_{t + 1} & = & \underbrace{\tilde{\theta}_t - \tilde{g}_t}_{\text{calculated in the
                 parameter server}}, \nonumber
\end{eqnarray}
where $r_t$ is the remainder of $t / c$.  $\tilde{\xi}_{t, s}$ denotes
the subscript of $f$ used to calculate the $s^{\tmop{th}}$ gradient in
the $t^{\tmop{th}}$ step in our new formulation. One can verify that
here $t$ increases $c$ times faster than the $i$ in \eqref{dl-eq:1}.

Note that the $\tilde{\tau}_t - r_t$ in \eqref{dl-eq:2} is always
positive\footnote{In \eqref{dl-eq:2} when a mini-batch is updated into
  the parameter, the counter ($t$) will increase by $1$, while in
  \eqref{dl-eq:1} the counter ($i$) increases by $1$ every $c$
  mini-batches updated into the parameter. For example if we have $3$
  mini-batches with staleness $1$ pushed, in \eqref{dl-eq:1} all
  $\tau_{i,l}$ will be $1$. However, in \eqref{dl-eq:2}, if one
  mini-batch is updated in iteration $t$, the staleness
  $\tilde{\tau}_{t+1}$ of the other two mini-batches becomes $2$, so
  we need to subtract the redundant part of the staleness caused by
  the difference in counting. Because the staleness after subtraction
  is exactly the original staleness in \eqref{dl-eq:1}, it is always
  positive.}. We use $\{ p_t \}_{t \in \mathbb{N}}$ to denote the difference
$p_t = \tilde{\tau}_t - r_t$. It immediately
follows that $p_t = \tau_{\lfloor t/c \rfloor, r_t}$.
From the Theorem 1 in \cite{liu-asgd-nips-2015} with some
modification, we have the following theorem, which indicates the
convergence rate and the linear speedup property of our algorithm.

\begin{theorem}
  Let $C_1,C_2,C_3,C_4$ be certain positive constants depending on the objective function $F(\theta)$. Under certain commonly used assumptions (please find in
  Theorem 1 in \cite{liu-asgd-nips-2015}), we can achieve an
  convergence rate of
  \begin{equation}
    \frac{1}{\sum_{t = 1}^T 1 / p_t}  \sum_{t = 1}^T \frac{1}{p_t} \mathbb{E}
    (\| \nabla F (\tilde{\theta}_t) \|^2) \leqslant 2 \frac{\sqrt{\frac{2C_1C_2}{\mu}\sum_{t = 1}^T
    \left( \frac{1}{p_t^2}  \right) }}{\sum_{t = 1}^T
    \frac{1}{p_t}}, \label{dl-eq:1-1}
  \end{equation}
  where $T$ is the total iteration number, if
  \begin{equation}
    \alpha_0 = \sqrt{\frac{C_1 c^2 \mu}{\sum_{t = 1}^T \left(
    \frac{2}{p_t^2} C_2 \right) }}, \label{dl-eq:3}
  \end{equation}
  under the prerequisite that
  \begin{equation}
    \alpha_0 \leqslant \frac{cC_2}{C_3 p_t \sum_{j = t - 2 n}^{t - 1}
    \frac{1}{p_j^2}},\quad \forall t, \label{dl-eq:5}
  \end{equation}
  and
  \begin{equation}
    C_3  \frac{\alpha_0}{cp_t} + C_4 n \frac{\alpha_0^2}{c^2 p_t} 
    \sum_{\kappa = 1}^{2 n} \frac{1}{p_{t + \kappa}} \leqslant 1,\quad \forall t
    \label{dl-eq:6}.
  \end{equation}
\end{theorem}

First note that \eqref{dl-eq:5} and \eqref{dl-eq:6} can always be satisfied
by selecting small enough $\alpha_0$ (or equivalently, large enough
$T$). Thus, if the learning rate is appropriately chosen in our
algorithm, the weighted average of the gradients (which is the LHS of
\eqref{dl-eq:1-1}) is guaranteed to converge. Also note that that to achieve this convergence rate, the batch size $\mu$
cannot be too large, since from \eqref{dl-eq:3} we can see that a
larger $\mu$ leads to a larger $\alpha_0$, which may not satisfy the
prerequisites \eqref{dl-eq:5} and \eqref{dl-eq:6}.

A clearer dependence between the staleness and the convergence rate
can be found by taking a closer look at the RHS \eqref{dl-eq:1-1}:

\begin{remark}
  Note that the RHS of \eqref{dl-eq:1-1} is of the form
  $h(z_1,\cdots,z_T)=O\left(\frac{\sqrt{z_1^2+z_2^2+\cdots+z_T^2}}{z_1+z_2+\cdots+z_T}\right)$
  by letting $z_t=\frac{1}{p_t}$. If the summation $z_1+\cdots+z_T$ is
  fixed, one can verify that $h$ is minimized when
  $z_1=z_1=\cdots=z_T$. Therefore our theorem suggests that a
  \emph{centralized} distribution of staleness $p$ (or $\tau$ in
  \eqref{dl-eq:1}) leads to a better convergence rate.
\end{remark}

Further, we have the following result by considering the ideal
scenario of the staleness.

\begin{remark}
  Note that if we take $p_t$ as a constant $p$, we have
  \[ \frac{1}{T} \sum_{t = 1}^T \mathbb{E} (\| \nabla F
    (\tilde{\theta}_t) \|^2) \leqslant 2 \frac{\sqrt{2 C_1 C_2
      }}{\sqrt{T{\mu}}} . \] Thus this convergence rate is roughly in
  the order of $O \left( 1 / \sqrt{\mu T} \right)$, where $T$ is the
  total iteration number and $\mu$ is the mini-batch
  size. Equivalently, a goal of
  $ \frac{1}{T} \sum_{t = 1}^T \mathbb{E} (\| \nabla F
  (\tilde{\theta}_t) \|^2)\le \epsilon$ can be achieved by having
  $\mu T=O(1/\epsilon^2)$. This is consistent with the convergence rate
  of SGD in \cite{liu-asgd-nips-2015}, which suggests a linear speedup
  can be achieved in our algorithm.
\end{remark}
\input{theory.tex}

%% file: theory.tex
\label{sec:proof}
\begin{proof}
  From \eqref{dl-eq:6} we have
  \begin{eqnarray*}
    &  & C_3  \frac{\alpha_0}{cp_t} + C_4 n \frac{\alpha_0^2}{c^2 p_t} 
    \sum_{\kappa = 1}^{2 n} \frac{1}{p_{t + \kappa}}\\
    & = & C_3 {\mu} \frac{\alpha_0}{{\mu}cp_t} + C_4 {\mu}^2 n
    \frac{\alpha_0}{{\mu}cp_t}  \sum_{\kappa = 1}^{2 n}
    \frac{\alpha_0}{{\mu}cp_{t + \kappa}}\\
    & \leqslant & 1, \forall t.
  \end{eqnarray*}
  With \eqref{dl-eq:5} we have
  \begin{equation}
    \frac{\alpha_0^2}{{\mu}c^2 p_t^2} C_2 \geqslant \frac{\alpha_0^3}{c^3
    p_t {\mu}} C_3  \sum_{j = t - 2 n}^{t - 1} \frac{1}{p_j^2}, \forall t.
    \label{dl-eq:7}
  \end{equation}
  Note that the upperbound of the staleness is $2 n$ in our
  setting. Then it follows from Theorem 1 in \cite{liu-asgd-nips-2015} that
  \begin{eqnarray*}
    &  & \frac{1}{\sum_{t = 1}^T 1 / p_t}  \sum_{t = 1}^T \frac{1}{p_t}
    \mathbb{E} (\| \nabla F (\tilde{\theta}_t) \|^2)\\
    & \leqslant & \frac{C_1 + \sum_{t = 1}^T \left(
    \frac{\alpha_0^2}{{\mu}c^2 p_t^2} C_2 + \frac{\alpha_0^3}{c^3 p_t
    {\mu}} C_3  \sum_{j = t - 2 n}^{t - 1} \frac{1}{p_j^2}
    \right)}{\sum_{t = 1}^T \frac{\alpha_0}{cp_t}}\\
    & \underbrace{\leqslant}_{\text{\eqref{dl-eq:7}}} & \frac{C_1 + \alpha_0^2 
    \sum_{t = 1}^T \left( \frac{2}{{\mu}c^2 p_t^2} C_2 \right) }{\sum_{t =
    1}^T \frac{\alpha_0}{cp_t}}\\
    & \underbrace{=}_{\text{\eqref{dl-eq:3}}} & 2 \frac{\sqrt{C_1 c \sum_{t =
    1}^T \left( \frac{2}{{\mu}cp_t^2} C_2 \right) }}{\sum_{t = 1}^T
    \frac{1}{p_t}}\\
    & = & 2 \frac{\sqrt{\frac{2C_1C_2}{\mu}\sum_{t =
    1}^T \left( \frac{1}{p_t^2} \right) }}{\sum_{t = 1}^T
    \frac{1}{p_t}},
  \end{eqnarray*}
  completing the proof.
\end{proof}

%% file: result.tex
\section{Experimental Results}
\label{sec:results}
\subsection{Hardware and Benchmark Datasets}
We deploy our implementation on a P775 supercomputer.
Each node of this system contains four eight-core \SI{3.84}{\giga \Hz} IBM POWER7 processors, one optical connect controller chip and \SI{128}{\giga \byte} of memory.
A single node has a theoretical floating point peak performance of \SI{982}{\giga flop/s}, memory bandwidth of \SI{512}{\giga \byte/s} and bi-directional interconnect bandwidth of \SI{192}{\giga\byte/\second}.

We present results on two datasets: \CIFAR and \imagenet.
The \CIFAR \cite{krizhevsky2009learning} dataset comprises of a total of 60,000 RGB images of size 32 $\times$ 32  pixels partitioned into the training set (50,000 images) and the test set (10,000 images). 
Each image belongs to one of the 10 classes, with 6000 images per class. 
For this dataset, we construct a deep convolutional neural network (CNN) with 3 convolutional layers each followed by a pooling layer. 
The output of the 3\ts{rd} pooling layer connects, via a fully-connected layer, to a 10-way softmax output layer that generates a probability distribution over the 10 output classes. 
This neural network architecture closely mimics the \CIFAR model available as a part of the open-source Caffe deep learning package (\cite{jia2014caffe}).
The total number of trainable parameters in this network are $\sim90$~K (model size of $\sim$\SI{350}{\kilo \byte}). The neural network is trained using momentum-accelerated mini-batch SGD with a batch size of 128 and momentum set to 0.9. As a data preprocessing step, the per-pixel mean is computed over the entire training dataset and subtracted from the input to the neural network. 

For \imagenet ~\cite{ILSVRC15}, we consider the image dataset used as a part of the 2012 \imagenet Large Scale Visual Recognition Challenge (ILSVRC 2012).
The training set is a subset of the \imagenet database and contains 1.2 million 256$\times$256 pixel images. The validation dataset has 50,000 images. 
Each image maps to one of the 1000 non-overlapping object categories. 
For this dataset, we consider the neural network architecture introduced in \cite{krizhevsky2012imagenet} consisting of 5 convolutional layers and 3 fully-connected layers. 
The last layer outputs the probability distribution over the 1000 object categories. In all, the neural network has $\sim$72 million trainable parameters and the total model size is \SI{289}{\mega \byte}. Similar to the \CIFAR benchmark, per-pixel mean computed over the entire training dataset is subtracted from the input image feeding into the neural network. 
\input{result_runtime.tex}

\input{result_accuracy.tex}

%% file: result_runtime.tex
\subsection{Runtime Evaluation}
\label{sec:result_runtime}
\begin{figure}
\vskip -0.2in
 \centering
	 \includegraphics[width=0.22\textwidth]
	 {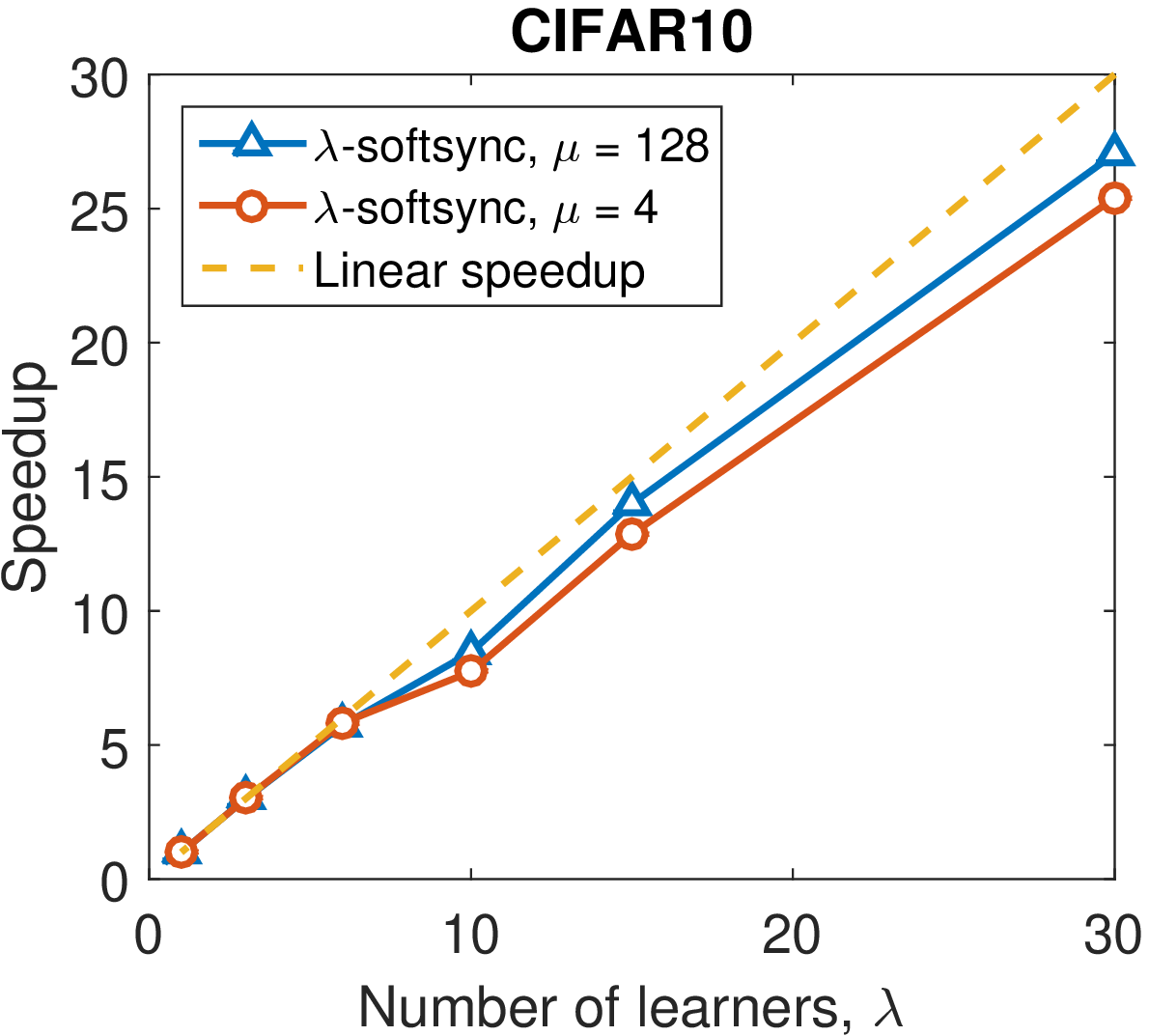}
\quad
	 \includegraphics[width=0.22\textwidth]
	 {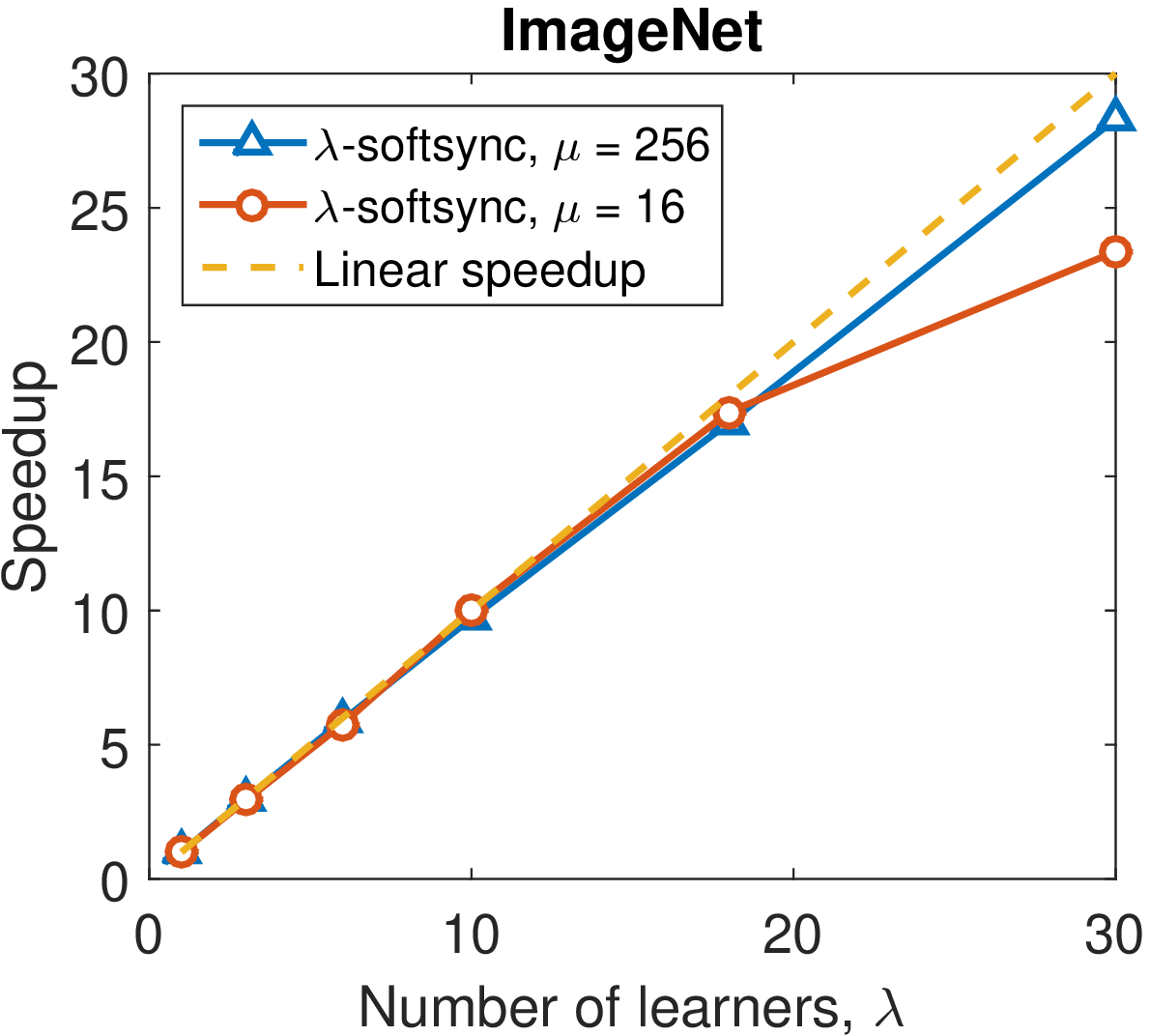}
	\caption{Measured speedup in training time per epoch for (a) \CIFAR(model size $\sim$\SI{350}{\kilo \byte}) and (b) \imagenet (model size $\sim$\SI{300}{\mega \byte})  }
        \label{fig:speedup}
\end{figure}
Figure ~\ref{fig:speedup} shows the speedup measured on CIFAR10 and ImageNet, for up to 30 learners. Our implementation achieves 22x-28x speedup for different benchmarks and different batch sizes. On an average, we find that the ASGD runs 50\% faster than its SSGD counterpart.

%% file: result_accuracy.tex
\subsection{Model Accuracy Evaluation}
\label{sec:result_accuracy}

For each of the benchmarks, we perform two sets of experiments: (a) setting learning rate fixed to the best-known learning rate for SSGD, $\alpha = \alpha_0$, and (b) tuning the learning rate on a per-gradient basis depending on the gradient staleness $\tau$,  $\alpha = \alpha_0/\tau$. It is important to note that when $\alpha = \alpha_0$ \textit{and} $n = \lambda$ (number of learners) in $n$-softsync protocol, our implementation is equivalent to the Downpour-SGD of \cite{distbelief}. Albeit at the cost of poor runtime performance, we also train using the hardsync protocol since it guarantees zero gradient staleness and achieves the best model accuracy. Model trained by Hardsync protocol provides the target model accuracy baseline for ASGD algorithm. Further, we perform distributed training of the neural networks for each of these tasks using the $n$-softsync protocol for \textit{different} values of $n$. This allows us to systematically observe the effect of stale gradients on the convergence properties.
 
\subsubsection{CIFAR10}
When using a single learner, the \minibatch size is set to 128 and training for 140 epochs using momentum accelerated SGD (momentum = 0.9) results in a model that achieves $\sim$18\% misclassification error rate on the test dataset. 
The base learning rate $\alpha_0$ is set to 0.001 and reduced by a factor of 10 after the 120\ts{th} and 130\ts{th} epoch. 
In order to achieve comparable model accuracy as the single-learner, we follow the prescription of~\cite{2015Gupta} and reduce the \minibatch size per learner as more learners are added to the system in order to keep the product of \minibatch size and number of learners approximately invariant. 

\begin{figure}[t]
\vskip -0.15in
\centering
	\includegraphics[width=0.5\textwidth ]{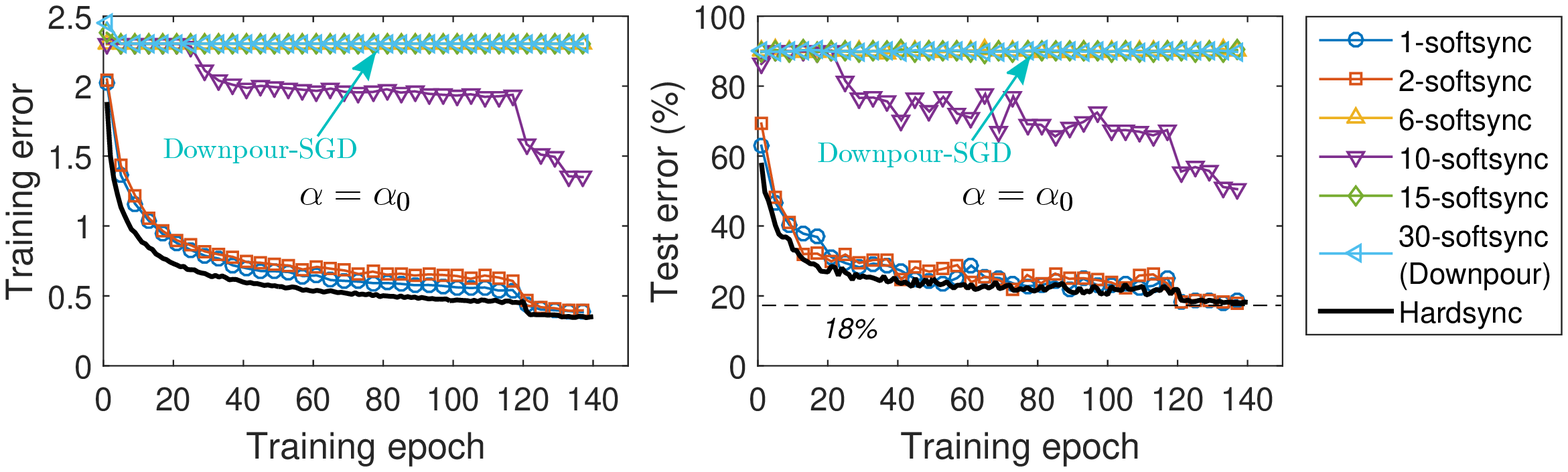}
	\label{fig:cifar10_control1}
\centering
	\includegraphics[width=0.5\textwidth]{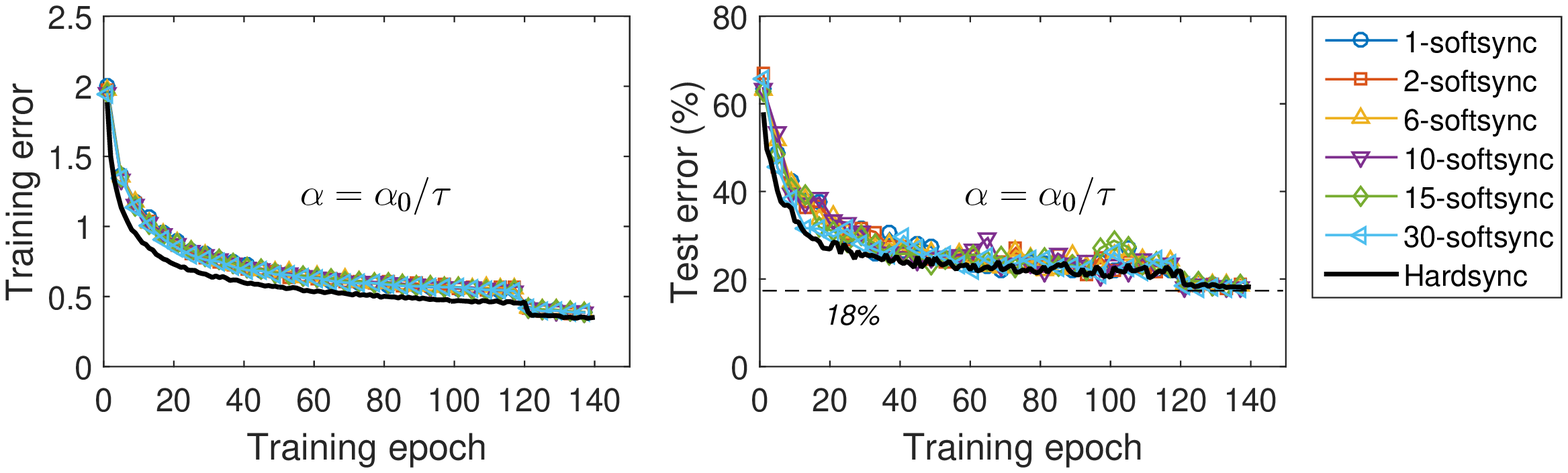}
\vskip -0.05in
\caption{\textit{\CIFAR results}: (a) Top: training error, test error for different $n$-softsync protocols, learning rate set as $\alpha_0$  (b) Bottom: staleness-dependent learning rate of Equation~\ref{eq:LR}. Hardsync (SSGD, black line), Downpour-SGD shown as baseline for comparison. $\lambda = 30$, $\mu = 4$. 
}
	\label{fig:cifar10_sps}
\end{figure}

Figure~\ref{fig:cifar10_sps} shows the training and test error obtained for different synchronization protocols: hardsync and $n$-softsync, $n \in (1,\lambda)$ when using $\lambda$ = 30 learners. The \minibatch size per learner is set to $4$ and all the other hyperparameters are kept unchanged from the single-learner case. Figure~\ref{fig:cifar10_sps} top half shows that as the gradient staleness is increased (achieved by increasing the splitting parameter $n$ in $n$-softsync protocol), there is a gradual degradation in SGD convergence and the resulting model quality. In the presence of large gradient staleness (such as in $15$, and $30$-softsync protocols), training fails to converge and the test error stays at 90\%. In contrast, Figure~\ref{fig:cifar10_sps} bottom half shows that when these experiments are repeated using our proposed staleness-dependent learning rate scheme of Equation~\ref{eq:LR}, the corresponding curves for training and test error for different $n$-softsync protocols are virtually indistinguishable (see Figure~\ref{fig:cifar10_sps} bottom half). Irrespective of the gradient staleness, the trained model achieves a test error of $\sim$18\%, showing that \emph{proposed learning rate modulation scheme is effective in bestowing upon the training algorithm a high degree of immunity to the effect of stale gradients}.

\subsubsection{ImageNet} 
With a single learner, training using a \minibatch size of 256, momentum 0.9 results in a top-1 error of 42.56\% and top-5 error of 19.18\% on the validation set at the end of 35 epochs. The initial learning rate $\alpha_0$ is set equal to 0.01 and reduced by a factor of 5 after the 20\ts{th} and again after the 30\ts{th} epoch. Next, we train the neural network using 18 learners, different $n$-softsync protocols and reduce the \minibatch size per learner to 16. 

\begin{figure}[tb]
\vskip -0.15in
\centering
	\includegraphics[width=0.5\textwidth]{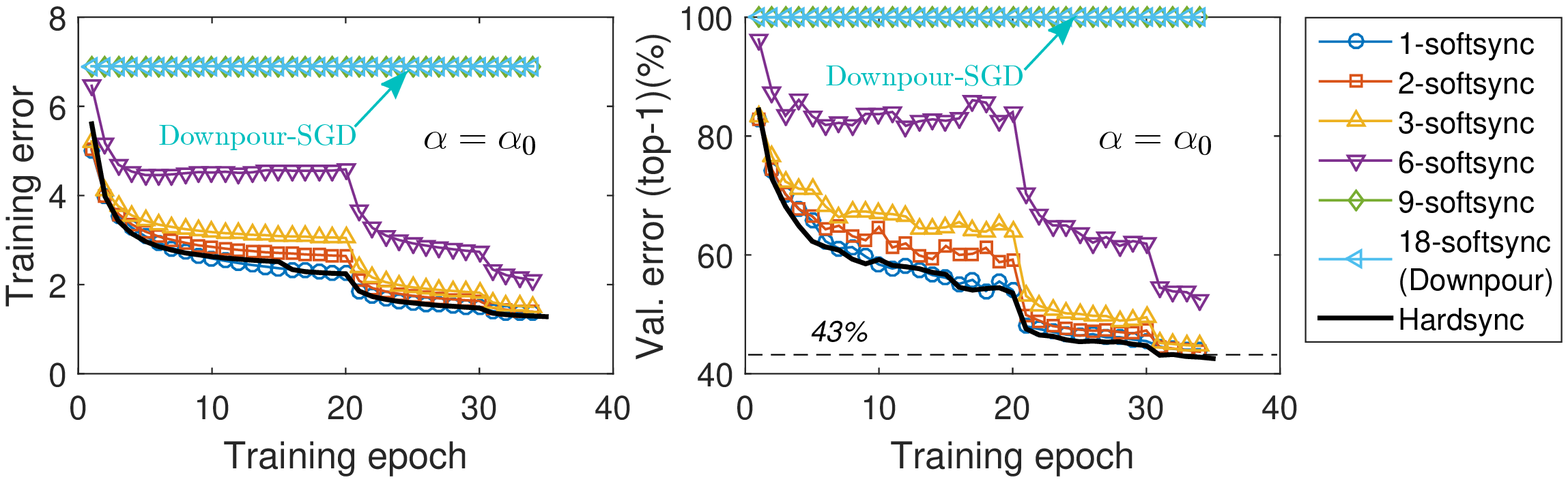}
	\label{fig:imagenet_control}
\centering
	\includegraphics[width=0.5\textwidth]{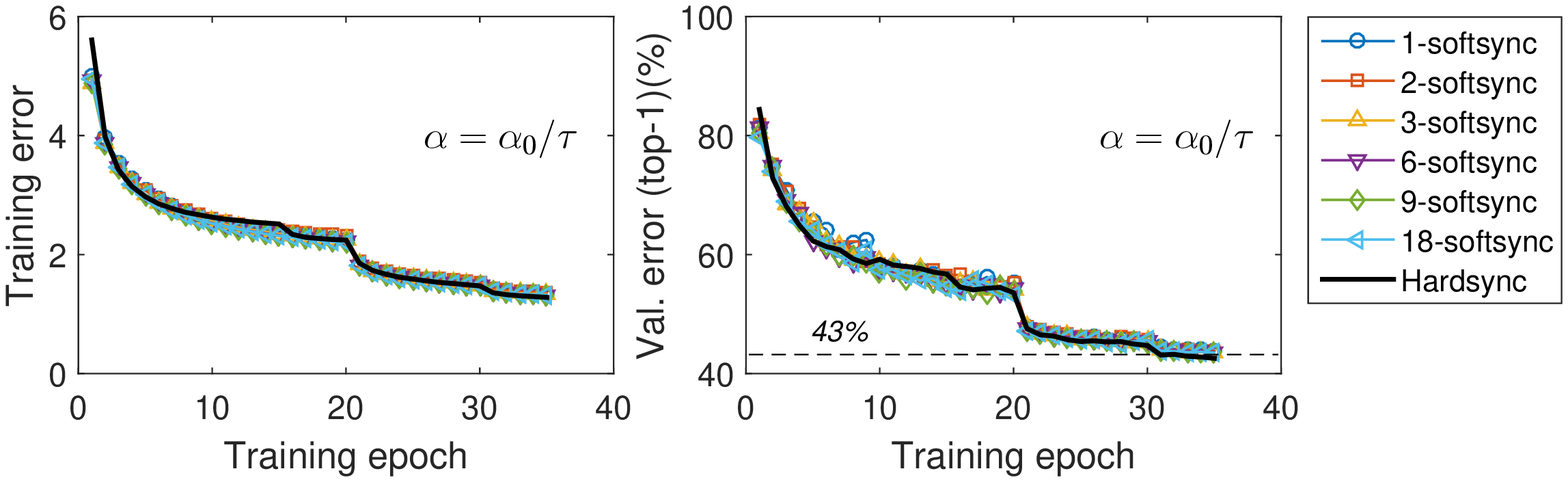}
\vskip -0.05in
\caption{\textit{\imagenet results}: (a) Top: training error, top-1 validation error for different $n$-softsync protocols, learning rate set as $\alpha_0$ (b) Bottom: staleness-dependent learning rate of Equation~\ref{eq:LR}. Hardsync(SSGD, black line), Downpour-SGD shown as baseline for comparison. $\lambda = 18$, $\mu = 16$. 
}
	\label{fig:imagenet_sps}
\end{figure}

Figure~\ref{fig:imagenet_sps} top half shows the training and top-1 validation error when using the learning rate that is the same as the single learner case $\alpha_0$. The convergence properties progressively deteriorate as the gradient staleness increases, failing to converge for $9$ and $18$-softsync protocols. 
On the other hand, as shown in Figure~\ref{fig:imagenet_sps} bottom half, automatically tuning the learning rate based on the staleness results in nearly identical behavior for all the different synchronization protocols.These results echo the earlier observation that the proposed learning rate strategy is effective in combating the adverse effects of stale gradient updates. Furthermore, adopting the staleness-dependent learning rate helps avoid the laborious manual effort of tuning the learning rate when performing distributed training using ASGD. 

\emph{\underline{Summary} With the knowledge of the initial learning rate for SSGD ($\alpha_0$), our proposed scheme can automatically tune the learning rate so that distributed training using ASGD can achieve accuracy comparable to SSGD while benefiting from near linear-speedup.} 

%% file: conclusion.tex
\section{Conclusion}
\label{sec:conclusion}
In this paper, we study how to effectively counter gradient staleness in a distributed implementation of the ASGD algorithm.
In summary, the key contributions of this work are:
\begin{itemize}
\item We prove that by using our proposed staleness-dependent learning rate scheme, ASGD can converge at the same rate as SSGD. 
\item We quantify the distribution of gradient staleness in our framework and demonstrate the effectiveness of the learning rate strategy on standard benchmarks (CIFAR10 and ImageNet). The experimental results show that our implementation achieves close to linear speedup for up to 30 learners while maintaining the same convergence rate in spite of the varying degree of staleness in the system and across vastly different data and model sizes.
\end{itemize}

\newpage